\documentclass[10pt,twocolumn,letterpaper]{article}

\usepackage{iccv}
\usepackage{times}
\usepackage{epsfig}
\usepackage{stackengine,graphicx}
\usepackage{amsmath}
\usepackage{amssymb}
\usepackage{array}
\usepackage{multirow}

\usepackage{booktabs,tabu}

\usepackage{color}

\DeclareGraphicsExtensions{.jpg,.png,.pdf}

\usepackage{nopageno}

\usepackage{tabularx}

\usepackage[pagebackref=true,breaklinks=true,letterpaper=true,colorlinks,bookmarks=false]{hyperref}

\iccvfinalcopy 

\def\iccvPaperID{1954} 

\ificcvfinal\fi
\begin{document}

\title{Learning High Dynamic Range from Outdoor Panoramas}

\author{Jinsong Zhang  \qquad Jean-Fran\c{c}ois Lalonde\\
Universit\'e Laval, Qu\'ebec, Canada\\
{\tt\small jinsong.zhang.1@ulaval.ca,  jflalonde@gel.ulaval.ca} \\
{\tt\small\url{http://www.jflalonde.ca/projects/learningHDR}}
}

\maketitle

\begin{abstract}
Outdoor lighting has extremely high dynamic range. This makes the process of capturing outdoor environment maps notoriously challenging since special equipment must be used. In this work, we propose an alternative approach. We first capture lighting with a regular, LDR omnidirectional camera, and aim to recover the HDR after the fact via a novel, learning-based inverse tonemapping method. We propose a deep autoencoder framework which regresses linear, high dynamic range data from non-linear, saturated, low dynamic range panoramas. We validate our method through a wide set of experiments on synthetic data, as well as on a novel dataset of real photographs with ground truth. Our approach finds applications in a variety of settings, ranging from outdoor light capture to image matching. 
\end{abstract}

\vspace{-1em}
\section{Introduction}

Outdoor lighting has an intrinsic dynamic range that is much higher than what conventional cameras can capture. While professional cameras boast dynamic ranges of up to 16 bits, we are still a long way from the full 22 bits needed to properly model outdoor lighting~\cite{stumpfel-afrigraph-04}. Therefore, to accurately capture the full dynamic range of outdoor illumination, one must resort to acquiring multiple exposures~\cite{debevec-siggraph-97}, imaging a specially-designed light probe~\cite{debevec-st-12}, or using custom-designed photographic hardware~\cite{manakov-siggraph-13,tocci-tog-11b}. 

An attractive alternative is to apply inverse lighting algorithms on low dynamic range imagery, which shifts the burden from capture to processing. These algorithms attempt to inverse the image formation process in order to recover lighting information, either in a physics-based~\cite{lombardi-eccv-12} or data-driven~\cite{lalonde-ijcv-12,hold-geoffroy-cvpr-17} way. A main limitation of these algorithms is that they are inherently limited to the information available within the image. An image may not always contain sufficient information to recover the lighting reliably. 

In this work, we propose to directly learn the relationship between the low dynamic range (LDR) information available in an outdoor $360^\circ$ panorama and the high dynamic range (HDR) of outdoor lighting. Our method takes as input a single LDR omnidirectional panorama, and converts it to HDR automatically, filling in saturated pixels with plausible values. Recovering HDR from LDR is known as inverse tonemapping: this is typically achieved by inverting the camera response function~\cite{rempel-siggraph-07}. While there exists a wide variety of such techniques in the literature, these methods are not tuned for outdoor lighting as they do not expect such extreme variations in dynamic range, and fail to recover plausible results as will be demonstrated in the paper. 

Our work proposes the following three key contributions. First, we present a full end-to-end learning approach which directly regresses HDR from LDR information in an outdoor panorama. Surely this is a challenging task: the sun can be 17 f-stops brighter than the rest of the sky~\cite{stumpfel-afrigraph-04}! To learn this relationship, we rely on a large set of HDR sky environment maps~\cite{hdrdb}, which we use as light sources to render a high quality synthetic city model to form a large corpus of synthetic panoramas. From this dataset, we train an LDR-to-HDR deep convolutional autoencoder, and show, through extensive experiments on synthetic and real data, that it succeeds in accurately predicting the extreme dynamic range of outdoor lighting. Our second key contribution is a novel dataset of real LDR panoramas and associated HDR ground truth. We use this novel dataset to see how the Convolutional Neural Network (CNN) can be adapted to work on the challenging case of real data. Our third key contribution is to show the applicability of our approach on three novel applications: single shot light probe, visualizing virtual objects and image matching in large LDR panorama database. Given the availability of large datasets of outdoor panoramas (e.g. SUN360~\cite{xiao2012recognizing} and Google Street View) and the recent interest in using HDR panoramas for virtual and mixed reality applications~\cite{rhee2017mr360}, our work is timely in enabling the use of LDR data, which is easy to capture, in HDR applications. Code and data are available on our project page.  

\vspace{-.5em}

\section{Related work}

\paragraph{Image-based lighting}
The seminal work of Debevec~\cite{debevec-siggraph-98} on image-based rendering demonstrated that capturing lighting can be achieved by acquiring several photographs of a mirrored sphere at different exposures. Since then, it has been demonstrated that the same can be done from a single shot of a metallic/diffuse hybrid sphere~\cite{debevec-st-12}. Similarly, specialized probes also exist for real-time applications~\cite{calian-sigtr-13}. Another method proposes to stitch multiple photographs taken with different exposures and viewpoints using a smartphone~\cite{kan-egshort-15}. The resulting light probes are useful for virtual object insertion, but also in virtual or mixed reality applications~\cite{rhee2017mr360}. In contrast, our approach uses a single LDR shot, taken from a commodity $360^\circ$ camera. 

\vspace{-1.1em}
\paragraph{Inverse tone mapping}
Algorithms for reproducing HDR from LDR images have been proposed in recent years, which are known as inverse tone mapping operators (iTMOs). To reproduce the HDR content from an LDR image is an ill-posed problem since the information is missing in the saturated regions. Banterle et al.~\cite{banterle-cgita-06, banterle2007framework} apply the inverse of the Reinhard tone mapping function~\cite{reinhard2002photographic} to the LDR image, then they create an \emph{expand map} by density estimation of the bright areas to guide the dynamic range expansion. Rempel et al.~\cite{rempel-siggraph-07} proposed a similar expand map combined with an edge stopping function to expand the dynamic range while increasing the contrast around edges. Kuo et al.~\cite{kuo2012content} use different inverse tone mapping parameters based on the scene content. Meylan et al.~\cite{meylan2006reproduction, meylan2007tone} explicitly focus on the specular highlight region; they use different linear functions to expand the diffuse and specular region in the image. The main concern of most of these techniques is to display LDR content onto HDR devices~\cite{akyuz2007hdr}, and as such these approaches are ill-suited for the case of outdoor lighting. 

\vspace{-1.1em}
\paragraph{Deep learning}
Deep CNNs are often used in image recognition and classification, but recent work has shown that they can also be used to estimate missing information from images. Pathak et al.~\cite{pathakCVPR16context} proposed to use CNNs to predict the missing content of a scene based on the surrounding pixels by minimizing a reconstruction loss and an adversarial loss~\cite{goodfellow2014generative}. Zhang et al.~\cite{zhang2016colorful} employed CNNs to recover color from grayscale images. Sangkloy et al.~\cite{sangkloy2016scribbler} proposed an adversarial image synthesis architecture to constrain the generation by user input.
CNNs are also used for generating a high resolution image from a low resolution image~\cite{denton2015deep}, predicting the missing depth information from a single RGB image~\cite{eigen2014depth}, producing a complete 3D voxel representation from a single depth image~\cite{song2016ssc}. In contrast, our approach recovers missing \emph{dynamic range}. 

\begin{figure}[!t] 
\centering
\includegraphics[width=\linewidth]{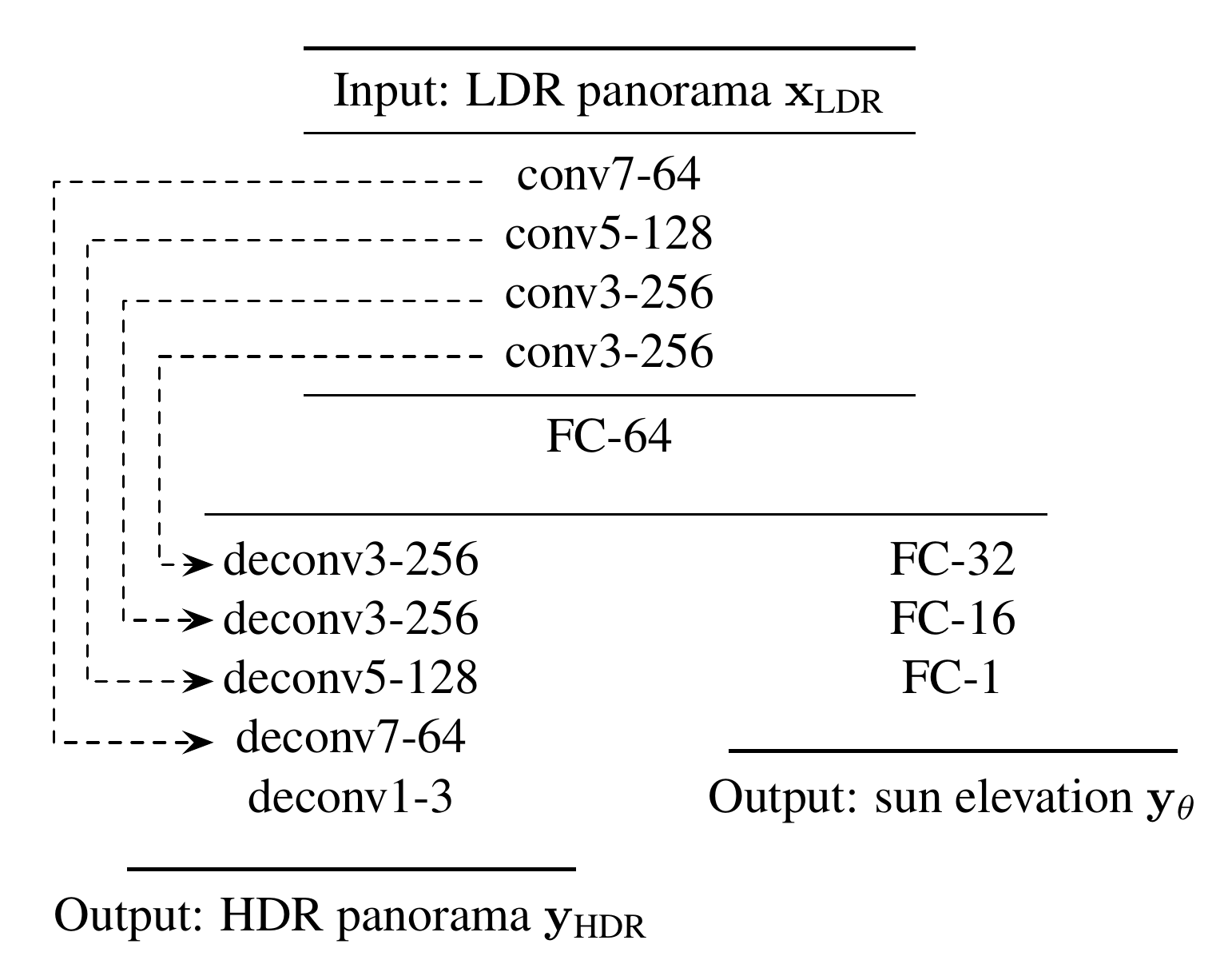}
\caption[]{The proposed CNN architecture. The encoder (top half) compresses the input LDR panorama to a 64-dimensional vector through 4 convolutional layers and splits into two heads: one to reconstruct the HDR panorama $\mathbf{y}_\text{HDR}$ through a series of deconvolutional layers; and the second, composed of two fully-connected layers, to predict the sun elevation $\mathbf{y}_\theta$. The output of each convolutional layer is added to the input of its deconvolutional counterpart via skip links (dashed lines). A stride of 2, batch normalization, and ELU~\cite{clevert-iclr-16} are used on all (de)convolutional layers.}
\label{fig:cnn-architecture}
\vspace{-1em}
\end{figure}

\vspace{-.2em}
\section{Approach}
\vspace{-.2em}
Our approach relies on a convolutional autoencoder that learns to reconstruct high dynamic range from low dynamic range panoramas. It is trained on a large dataset of synthetic LDR panoramas. Before the data generation process is described in sec.~\ref{sec:validation-synthetic}, we first describe our CNN architecture, loss function, and training parameters. 

\subsection{Network architecture}

As shown in fig.~\ref{fig:cnn-architecture}, we use a convolutional neural network that takes in a $64 \times 128$ LDR panorama $\mathbf{x}_\text{LDR}$ as input stored in the latitude-longitude format. It assumes the panorama has been previously rotated to align the sun with the center of the image\footnote{As in~\cite{hold-geoffroy-cvpr-17}, the sun is detected by computing the center of mass of the largest saturated region in the image.}. From this input, it produces a 64-dimensional encoding of the input through a series of convolutions downstream and splits into two upstream expansions, with two distinct tasks: (1) HDR recovery, and (2) sun elevation prediction. The encoder and decoder are both modeled by four (de)convolution layers. The output of each convolution is added to the input of its corresponding deconvolution layer via skip links (shown with dashed lines in fig.~\ref{fig:cnn-architecture}). Each (de)convolution layer is followed by an (up-sub)sampling step (stride of 2), batch normalization, and the ELU activation function~\cite{clevert-iclr-16}. The sun elevation decoder is composed of three low-dimensional fully-connected layers. 

\begin{figure}[!t]
\footnotesize
\centering
\begin{tabular}{cc}
\includegraphics[height=2.8cm,trim={0 3cm 0 0},clip]{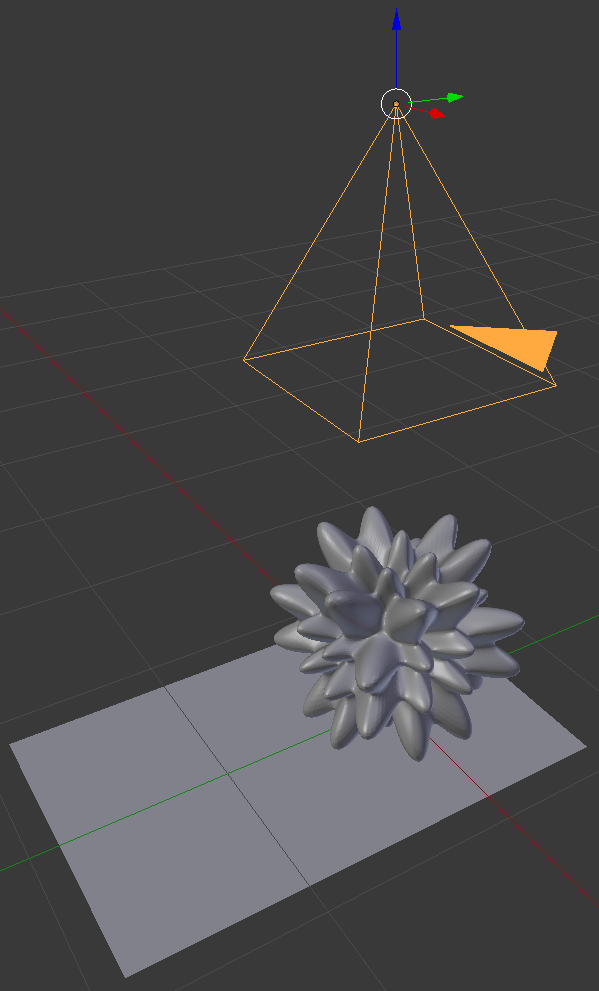} &
\includegraphics[height=2.8 cm]{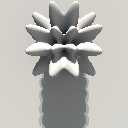} \\
(a) 3D scene & (b) Example rendering
\end{tabular}
\vspace{.5em}
\caption{(a) 3D view of the scene used to compute the transport matrix $\mathbf{T}$ in the rendering loss $\mathcal{L}_\text{render}$ (\ref{eqn:loss-render}), and (b) example rendering obtained with the resulting $\mathbf{T}$ and a sunny HDR panorama. The perspective effect is considerable when the camera is close to the object, as such, the shadow cast on the plane is smaller than the object. The scene was chosen to generate complex lighting effects such as cast shadows, smooth shading, and occlusions.}
\label{fig:spiky-sphere}
\vspace{-1em}
\end{figure}

\subsection{Loss function}

To train the CNN, the following loss function is defined: 
\begin{equation}
\mathcal{L}_\text{all}(\mathbf{y}, \mathbf{t}) = 
\mathcal{L}_\text{HDR}(\mathbf{y}, \mathbf{t}) + 
\lambda_{1}\mathcal{L}_\theta(\mathbf{y}, \mathbf{t}) + 
\lambda_{2}\mathcal{L}_\text{render}(\mathbf{y}, \mathbf{t}) \,,
\label{eqn:loss-all}
\end{equation}
where $\mathbf{y}$ is the predicted output, and $\mathbf{t}$ is the ground truth target label. Eq. (\ref{eqn:loss-all}) is composed of three loss functions computed on different labelled data, with the weights $\lambda_{1} = \lambda_{2} = 0.1$ controlling the relative importance of each loss function. 
The first element in (\ref{eqn:loss-all}) is the L1 norm between the predicted HDR panorama $\mathbf{y}_\text{HDR}$ and the ground truth $\mathbf{t}_\text{HDR}$: 
\begin{equation}
\mathcal{L}_\text{HDR}(\mathbf{y}, \mathbf{t}) = || \mathbf{y}_\text{HDR} - \mathbf{t}_\text{HDR} ||_1 \,.
\end{equation}	
The L1 norm is used to be more robust to the very high dynamic range of the HDR panorama. Since the sun pixel intensity can be up to 100,000 times brighter than other pixels, using an L2 norm overwhelmingly penalizes errors on the sun at the expense of the other, lower dynamic range pixels. To help the network in predicting high dynamic range values, the full HDR target $\mathbf{t}^*_\text{HDR}$ is tonemapped using
\begin{equation}
\mathbf{t}_\text{HDR} = \alpha (\mathbf{t}^*_\text{HDR})^{1/\gamma} \,.
\label{eqn:loss-tonemapping}
\end{equation}	
We use $\gamma=2.2$ and $\alpha=1/30$, which brings the sun intensity close to 1 when it is bright. The inverse of (\ref{eqn:loss-tonemapping}) is applied to $\mathbf{y}_\text{HDR}$ to convert the network output to full HDR. 

The second element in (\ref{eqn:loss-all}) computes the L2 norm between the predicted and ground truth sun elevations $\mathbf{y}_\theta$, $\mathbf{t}_\theta$: 
\begin{equation}
\mathcal{L}_\theta(\mathbf{y}, \mathbf{t}) = || \mathbf{y}_\theta - \mathbf{t}_\theta ||_2 \,.
\end{equation}	
Our experiments demonstrate that a small gain in performance can be obtained with this additional path in the network (see sec.~\ref{sec:synthetic-quantitative}). Finally, we also incorporate a render loss in the third element in (\ref{eqn:loss-all}):
\begin{equation}
\mathcal{L}_\text{render}(\mathbf{y}, \mathbf{t}) = || \mathbf{T}\mathbf{y}_\text{HDR} - \mathbf{T}\mathbf{t}_\text{HDR} ||_2 \,,
\label{eqn:loss-render}
\end{equation}	
where $\mathbf{T}$ is the transport matrix for a lambertian scene (without interreflections). To ensure that interesting lighting effects are captured, a scene made of a complex ``spiky sphere'' on a flat ground plane seen from above is used (see fig.~\ref{fig:spiky-sphere}). This effectively re-weights the pixels in the panoramas according to the fraction of the visible hemisphere for each pixel in the scene. Rendering is performed at $64 \times 64$ resolution, so $\mathbf{T}$ is of dimensions\footnote{The number of columns in $\mathbf{T}$ is equal to the number of pixels in the panorama, so $64 \times 128 = 8,192$. To save memory, it can be divided by 2 because only the top hemisphere is visible in the render.} $64^2 \times 4,096$. Since this is a simple matrix multiplication, gradients can easily be back-propagated through this rendering step. 

\subsection{Training details}

To train the CNN, we use the ADAM optimizer~\cite{kingma-iclr-15} with a minibatch size of $128$, initial learning rate of $0.001$, and momentum parameter of $\beta_1=0.9$. Training 500 epochs takes roughly 50 hours on an Nvidia GeForce 1060 GPU. At test time, inference takes approximately 5ms. 

\begin{figure}[!t]
\centering
\includegraphics[width=.9\linewidth]{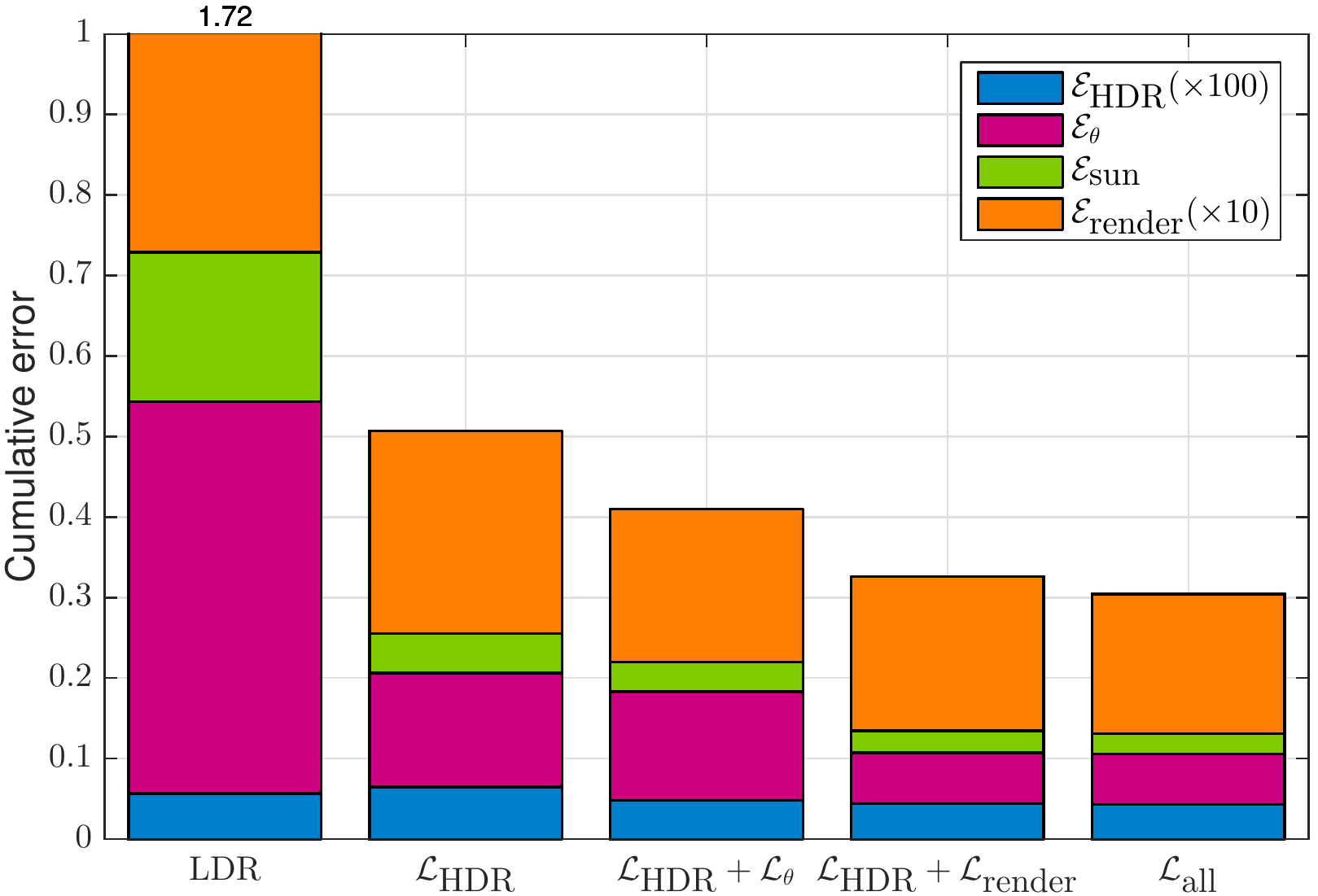}
\caption{Quantitative comparison of results obtained with the LDR panorama (leftmost column) and different combinations of losses (other columns) for our approach on the synthetic test set. Different error metrics are reported, from top to bottom: the mean absolute error on the sky panorama $\mathcal{E}_\text{HDR}$, the RMSE on the sun elevation $\mathcal{E}_\theta$, the RMSE on the sun intensity $\mathcal{E}_\text{sun}$, and the RMSE on the render $\mathcal{E}_\text{render}$. $\mathcal{L}_\text{all}$ yields the lowest error on all metrics. }
\label{fig:synthetic-quantitative} 
\vspace{-1em}
\end{figure}

\begin{figure*}[!t]
\centering
\footnotesize
\setlength{\tabcolsep}{1pt}
\newcolumntype{R}{c@{\extracolsep{1pt}}c@{\extracolsep{5pt}}}%
\begin{tabular}{cRRR}
& Panorama & Render & Panorama & Render & Panorama & Render \\
\rotatebox{90}{\hspace{.1em}Gnd truth HDR}& 
\includegraphics[height=1.82cm]{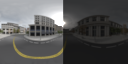} &
\includegraphics[height=1.82cm]{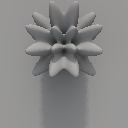} &
\includegraphics[height=1.82cm]{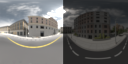} &
\includegraphics[height=1.82cm]{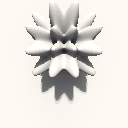} &
\includegraphics[height=1.82cm]{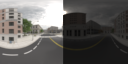} &
\includegraphics[height=1.82cm]{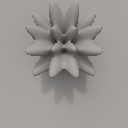} \\
\rotatebox{90}{\hspace{.1em}LDR panorama}& 
\includegraphics[height=1.82cm]{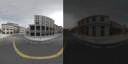} &
\includegraphics[height=1.82cm]{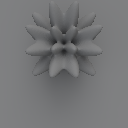} &
\includegraphics[height=1.82cm]{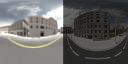} &
\includegraphics[height=1.82cm]{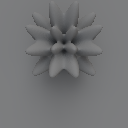} &
\includegraphics[height=1.82cm]{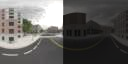} &
\includegraphics[height=1.82cm]{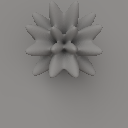} \\
\rotatebox{90}{\hspace{.1em}Predicted HDR}& 
\includegraphics[height=1.82cm]{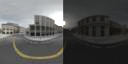} &
\includegraphics[height=1.82cm]{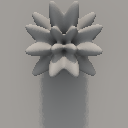} &
\includegraphics[height=1.82cm]{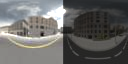} &
\includegraphics[height=1.82cm]{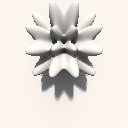} &
\includegraphics[height=1.82cm]{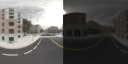} &
\includegraphics[height=1.82cm]{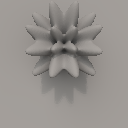} 
\end{tabular}
\caption{Qualitative results on the synthetic dataset. \emph{Top row}: the ground truth HDR panorama, \emph{middle row}: the LDR panorama, and \emph{bottom row}: the predicted HDR panorama obtained with our method. To illustrate dynamic range, each panorama is shown at two exposures, with a factor of 16 between the two. For each example, we show the panorama itself (left column), and the rendering of a 3D object lit with the panorama (right column). The object is a ``spiky sphere'' on a ground plane, seen from above. Our method accurately predicts the extremely high dynamic range of outdoor lighting in a wide variety of lighting conditions. A tonemapping of $\gamma=2.2$ is used for display purposes. Please see additional examples on our project page. }
\label{fig:synthetic-qualitative}
\vspace{-1em}
\end{figure*}

\begin{figure}[!t]
\centering
\footnotesize
\includegraphics[width=\linewidth]{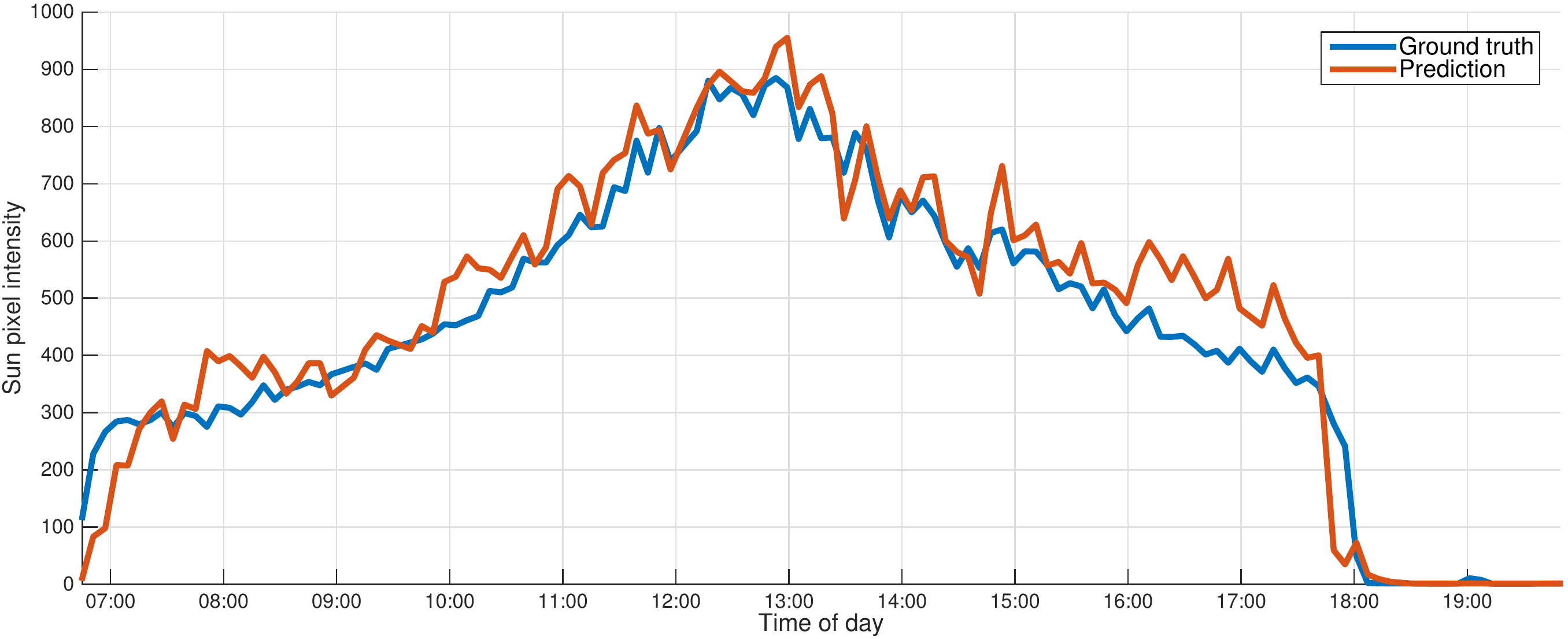}
\caption{HDR predictions over the course of one day from the synthetic test set. We plot the ground truth sun intensity (blue), along with the predictions from our network (orange) over the course of one day. Our method achieves a relatively stable temporal coherence, even if it is working on one panorama at a time. }
\label{fig:synthetic-time}
\vspace{-2em}
\end{figure}


\section{Validation on Synthetic Data}
\label{sec:validation-synthetic}

\subsection{Data generation}

Even though small datasets of HDR panoramas exist\footnote{See for example: \url{http://www.hdrlabs.com}.}, none do in sufficiently large quantity to train a deep neural network. Therefore, we have obtained access to the Laval HDR Sky Database~\cite{hdrdb}, which contains approximately 38,000 unsaturated, HDR omnidirectional photos of the sky, gathered over 103 different days over the course of 3 years. We use a subset of 9,732 HDR skies to generate renders of a realistic virtual 3D model of a small city. The 3D model was obtained from the Unity Store, and contains over 100 modular buildings with different styles and materials, including realistic roads, sidewalks, and foliage. 

To generate a panorama, the sky is first rotated to center the sun in the middle of the panorama based on the known ground truth sun position. Next, a random camera position is sampled in manually defined regions in the 3D city model corresponding to open spaces. The HDR sky is then set as the sole virtual light source, and rendering is performed. For data augmentation, each HDR sky is flipped horizontally and re-exposed with factors $1.75^x$, where $x=\{-1,0,1\}$. We employ the physically-based rendering engine Mitsuba~\cite{Mitsuba}, and render with a virtual omnidirectional camera, saving the output directly to a latitude-longitude panorama format in EXR. The LDR panorama $\mathbf{x}_\text{LDR}$ is obtained by converting this output to an 8-bit JPG file. Example images generated by this process are shown in fig.~\ref{fig:synthetic-qualitative}. 

From the resulting dataset of synthetic panoramas, $69\%$ (70 days, 39,198 images) of the samples are used for training, $15\%$ (16 days 8,730 images) for validation and early stopping, and $16\%$ (17 days, 10,458 images) for test. Note that care is taken to split the dataset according to \emph{days}, since HDR sky images are taken at the frequency of one every two minutes in \cite{hdrdb}, so two consecutive photos from the same day are extremely similar. 

\subsection{Quantitative experiments}
\label{sec:synthetic-quantitative} 

The model is first evaluated on the synthetic test set. Different loss functions are compared together in fig.~\ref{fig:synthetic-quantitative}. We evaluate the performance using different error metrics: the mean absolute error (MAE) on the HDR sky $\mathcal{E}_\text{HDR}$, RMSE on the sun elevation $\mathcal{E}_\theta$, and RMSE on the ``spiky sphere'' render $\mathcal{E}_\text{render}$ (fig.~\ref{fig:spiky-sphere}). We also compute the RMSE on the predicted sun intensity $\mathcal{E}_\text{sun}$, which is approximated by the intensity of the brightest pixel in the HDR image.

Fig.~\ref{fig:synthetic-quantitative} shows that the model trained solely with $\mathcal{L}_\text{HDR}$ already provides a good result when compared to using the LDR panorama directly. While various combinations of losses improve upon the baseline, combining the three losses in $\mathcal{L}_\text{all}$ (\ref{eqn:loss-all}) yields the lowest error on all metrics. Fig.~\ref{fig:synthetic-qualitative} shows qualitative results obtained by the network. Interestingly, predicting the sun elevation and estimating the HDR benefit from each other, as including them in the loss function results in improved performance at both these tasks.

\subsection{Temporal coherence}

We select one full day from sunrise to sunset in the HDR sky dataset~\cite{hdrdb}, and relight the city model from a fixed camera position using that day. Our model is used to regress the HDR from the resulting LDR panoramas one at a time. Fig.~\ref{fig:synthetic-time} shows that even if we do not enforce temporal consistency, our network successfully adapts to time changes and corresponding variations in sun intensity, as the prediction closely follows the ground truth. 

\section{Experiments on real data} 

We now present extensive experiments which validate that our approach is applicable to real-world data. First, we present a novel dataset of real outdoor LDR panoramas with corresponding HDR ground truth. This dataset is used first in a quantitative comparison of several training approaches; and second in a comparison to previous work. 

\subsection{Dataset of real photographs}

We collected a novel dataset of LDR panoramas with their corresponding HDR skies. The LDR panoramas were captured with a Ricoh Theta S camera, a consumer grade point-and-shoot $360^\circ$ camera. The HDR skies were captured with a Canon 5D Mark iii mounted on a tripod, equipped with a Sigma 8mm fisheye lens, and placed at the same location as the Theta camera. To properly image the true dynamic range of outdoor lighting, we installed a ND 3.0 filter behind the lens and captured 7 exposures ranging from $1/8000$ to 8 seconds at f/16, following~\cite{stumpfel-afrigraph-04}. The fisheye lens was radiometrically and geometrically calibrated~\cite{scaramuzza-iros-06}, so that the resulting HDR image could be warped to a latitude-longitude panorama. 

We use a ColorChecker Digital SG chart to match the colors between the LDR and HDR panoramas. To compensate for possible misalignments, the two panoramas are first geometrically aligned by finding SIFT feature correspondences and using RANSAC to find the optimal rotation matrix between the panoramas. This procedure is followed by a step of pixel-wise optical flow refinement to account for lens calibration errors. 

\begin{figure}[!t]
\footnotesize
\centering 
\includegraphics[width=.49\linewidth]{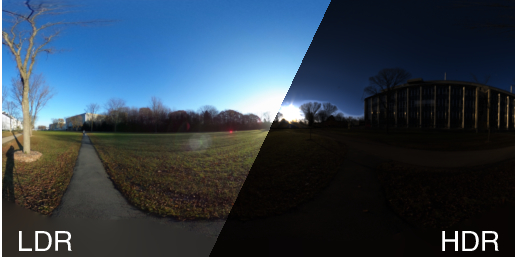} 
\includegraphics[width=.49\linewidth]{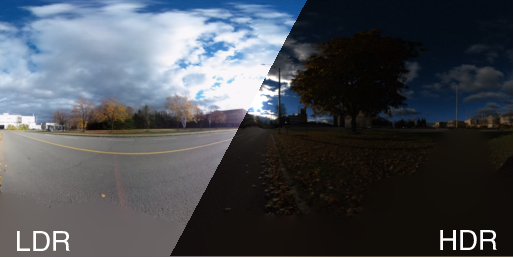} 
\includegraphics[width=.49\linewidth]{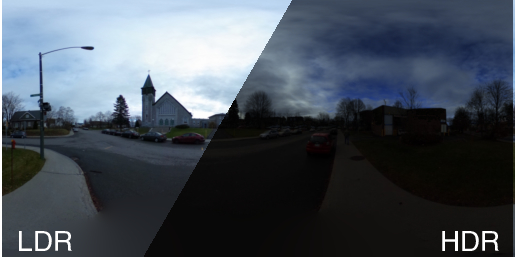}
\includegraphics[width=.49\linewidth]{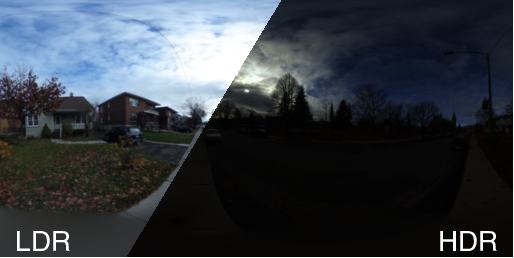} 
\caption{Examples from our real dataset. For each case, we show the LDR panorama captured by the Ricoh Theta S (left), and the corresponding HDR panorama captured by the Canon 5D Mark iii (right, shown at a different exposure to illustrate the high dynamic range). Please see additional examples on our project page. }
\label{fig:real-dataset}
\vspace{-1.5em}
\end{figure}

Using this technique, we captured pairs of real LDR/HDR panoramas over 13 different days, for a total of 404 pairs. Our dataset contains a variety of different illumination and weather conditions, as shown in fig.~\ref{fig:real-dataset}. The dataset is separated into non-overlapping training and test subsets of 8 and 5 days (490 and 318 panoramas---obtained by flipping each panorama horizontally) respectively. The same split is used in the following experiments to allow for comparisons between techniques. Please see additional results on our project page.

\begin{figure*}[!t]
\centering
\footnotesize
\setlength{\tabcolsep}{1pt}
\begin{tabular}{cc}
\includegraphics[width=.45\linewidth, trim={0 1cm 0 0}, clip]{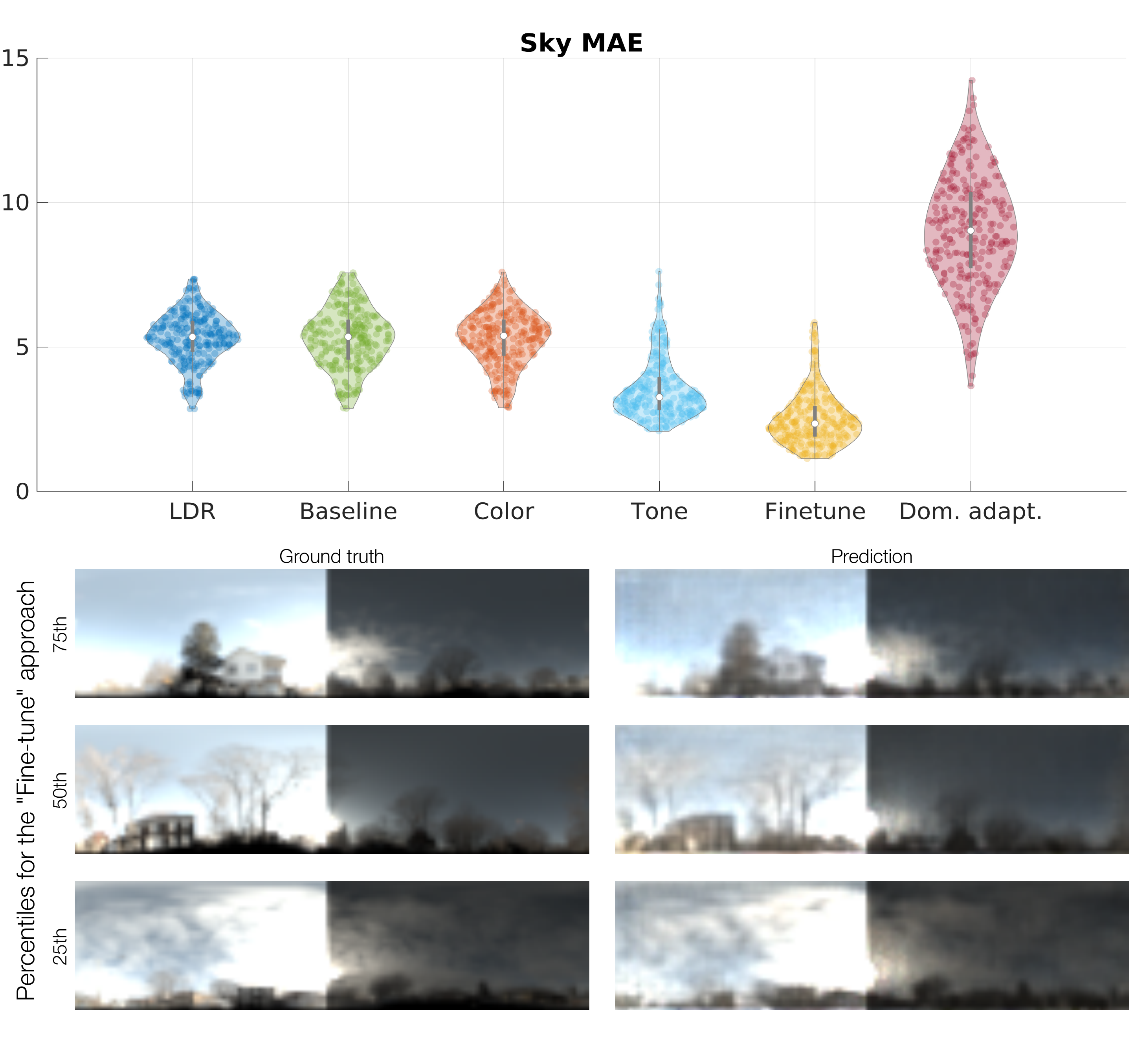} & 
\includegraphics[width=.45\linewidth, trim={0 1cm 0 0}, clip]{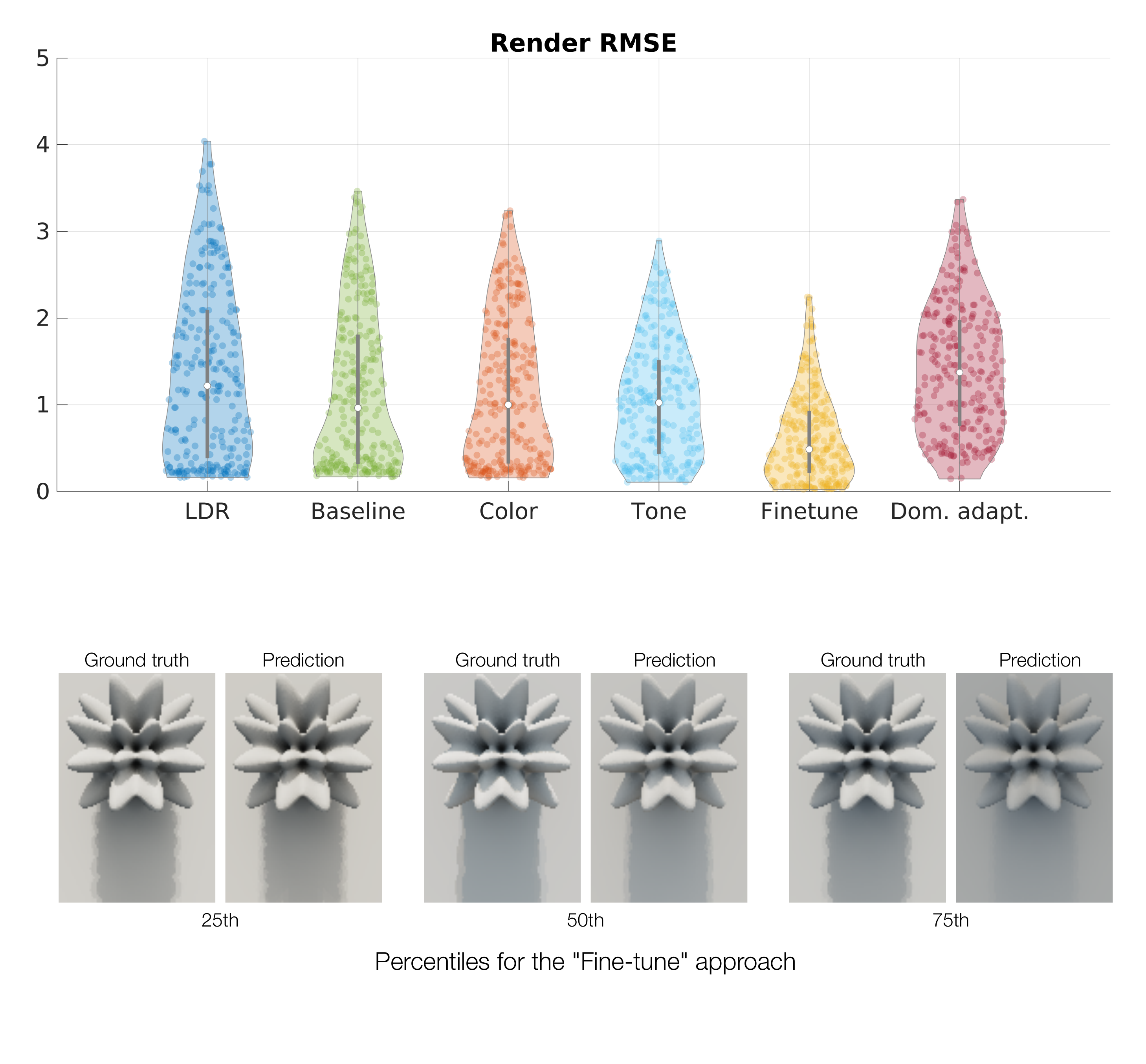} \\
(a) MAE on the sky & (b) RMSE on the render
\end{tabular}
\vspace{.5em}
\caption{Quantitative comparison of models on the real dataset. For each model, we show a full distribution of errors (curved shapes), as well as the 25th, 50th and 75th percentiles (gray vertical bar). }
\label{fig:quant-violin} 
\vspace{-2em}
\end{figure*}

\subsection{Adapting to real data}
\label{sec:quantitative-evaluation}

\begin{table}[!t]
\centering
\begin{tabular}{lcccc}
\toprule
Method & $\mathcal{E}_\text{HDR}$ & $\mathcal{E}_\text{render}$ & $\mathcal{E}_\theta$ & $\mathcal{E}_{\text{sun}}$ \\
\midrule
LDR  			& 5.30 & 1.34 & 0.21 & 0.54 \\
Baseline	 	& 5.34 & 1.19 & 0.10 & 0.43 \\
Color 			& 5.32 & 1.16 & 0.10 & 0.46 \\
Tone 			& \textbf{3.59} & \textbf{1.06} & \textbf{0.08} & \textbf{0.31} \\
Domain adaptation~\cite{ganin-icml-15}		& 7.53 & 1.30 & 0.10 & 0.36 \\
\midrule
Fine-tuning		& \textbf{2.55} & \textbf{0.64} & \textbf{0.07} & \textbf{0.22} \\
\bottomrule
\end{tabular}
\vspace{.5em}
\caption{Analyzing the impact of training data.
 Quantitative comparison of models trained on different synthetic inputs, and tested on real JPG panoramas. Refer to fig.~\ref{fig:quant-violin} for qualitative examples. The model is trained on: linear data (``Baseline''), linear data with color changes (``Color''), and non-linear data where a camera response function is applied along with color changes (``Tone''). The network trained on the ``Tone'' data achieves better performance when testing on real panoramas. Domain adaptation~\cite{ganin-icml-15} does not perform well on the real dataset; we believe this is because we do not have enough Theta S images. Fine-tuning with the ground truth training data (``Fine-tuning'') performs best, see sec.~\ref{subsubsection-finetuning}.} 

\label{tab:quant-augmentation}
\end{table}

\subsubsection{Augmenting the training dataset}

Evaluating the network trained on linear data on the real world data yields the performance shown in the first row of table~\ref{tab:quant-augmentation}. To improve these results, we first attempt to augment the synthetic dataset used for training, in order to more closely match real world capture conditions.

\begin{table}[!t]
\centering
\begin{tabular}{clcccc}
\toprule
& Method & $\mathcal{E}_\text{HDR}$ & $\mathcal{E}_\text{render}$ & $\mathcal{E}_\theta$ & $\mathcal{E}_{\text{sun}}$ \\
\midrule
\multirow{3}{*}{\rotatebox{90}{JPGs}}
& LDR  			& 5.30 & 1.34 & 0.21 & 0.54 \\
& Baseline	 	& 5.34 & 1.19 & 0.10 & 0.43 \\
& Tone 			& 3.59 & 1.06 & 0.08 & 0.31 \\
\midrule
\multirow{3}{*}{\rotatebox{90}{$\gamma=2.2$}}
& LDR  			& 6.33 & 1.53 & 0.19 & 0.54 \\
& Baseline	 	& 6.63 & 1.45 & 0.13 & 0.49 \\
& Tone 			& 7.06 & 1.50 & 0.08 & 0.43 \\
\midrule
\multirow{3}{*}{\rotatebox{90}{RF}}
& LDR  			& 4.64 & 1.46 & 0.18 & 0.54 \\
& Baseline	 	& 4.33 & 1.18 & 0.08 & 0.36 \\
& Tone 			& 6.64 & 1.35 & \textbf{0.07} & 0.35 \\
\midrule
\multirow{3}{*}{\rotatebox{90}{RF+WB}}
& LDR  			& 3.31 & 1.34 & 0.23 & 0.54 \\
& Baseline	 	& \textbf{2.99} & \textbf{1.03} & 0.08 & \textbf{0.30} \\
& Tone 			& 6.55 & 1.41 & 0.08 & 0.38 \\
\bottomrule
\end{tabular}
\vspace{.5em}
\caption{Analyzing the impact of modeling the camera. Radiometric camera models of increasing fidelity are applied to the JPG inputs and compared. Camera models include: none (``JPGs''), $\gamma = 2.2$, calibrated inverse response function (``RF''), and combining the ``RF'' with white balance adjustment (``RF+WB''). The three methods compared are: using the input LDR directly (``LDR''), network trained on synthetic linear data (``Baseline''), and network trained on synthetic data augmented with color shifts and non-linear response functions (``Tone'', see also table~\ref{tab:quant-augmentation}). }
\label{tab:quant-linearization}
\vspace{-1em}
\end{table}

\begin{table}[!t]
\centering
\begin{tabular}{lcccc}
\toprule
Method & $\mathcal{E}_\text{HDR}$ & $\mathcal{E}_\text{render}$ & $\mathcal{E}_\theta$ & $\mathcal{E}_{\text{sun}}$ \\
\midrule
LDR  							& 5.30 & 1.34 & 0.21 & 0.54 \\
\cite{banterle2007framework} 	& 16.5 & 9.24 & 0.16 & 0.30 \\
\cite{kuo2012content}			& 24.63 & 19.30 & 0.18 & 0.30 \\
\cite{meylan2006reproduction}	& 29.1 & 29.25 & 0.17 & 0.28 \\
\cite{rempel-siggraph-07}		& 29.1 & 29.26 & 0.17 & 0.29 \\
\cite{hold-geoffroy-cvpr-17}	& 5.4  & 21.8 & \textbf{0.06} & 1.70 \\
Ours							& \textbf{2.55} & \textbf{0.64} & 0.07 & \textbf{0.22} \\
\bottomrule
\end{tabular}
\vspace{.5em}
\caption{Comparison with previous work. The main parameters of each competing method are cross-validated on the ground truth training data, while our approach is fine-tuned using the same data. The existing inverse tonemapping methods yield unreasonable estimates when regressing the extreme HDR of outdoor lighting.}
\label{tab:quant-comparison}
\vspace{-1.5em}
\end{table}

\vspace{-1em}
\paragraph{Modeling the white balance} Digital cameras often apply significant color adjustments to images. We simulate this by applying a random additive shift to the hue and saturation channels $(s_h,s_s)$, where $s_h \sim \mathcal{N}(0, 10)$ and $s_s \sim \mathcal{N}(0, 0.1)$. Adding color shifts to the training data only marginally improves performance (table~\ref{tab:quant-augmentation}).

\vspace{-1em}
\paragraph{Modeling the response function of the camera} 
Real cameras have non-linear response functions. To simulate this, we randomly sample real camera response functions from the Database of Response Functions (DoRF)~\cite{grossberg2003space}, and apply them to the linear synthetic data before training. Training on this data yields a network which performs significantly better than the previous versions (see table~\ref{tab:quant-augmentation}). This also has the interesting side effect of resulting in better sun elevations. 

\vspace{-1em}
\paragraph{Domain adaptation with unlabelled data}
We apply unsupervised domain adaptation~\cite{ganin-icml-15} to adapt the synthetic model to real world images. This is achieved by adding a domain classifier connected to the FC-64 latent layer of the network via a \emph{gradient reversal layer}, which ensures the encoding feature is domain-invariant to synthetic and real data. The discriminator contains two fully-connected layers of 32 and 2 nodes, followed by ELU and softmax activation respectively. This model is trained by adding unlabelled real data to each minibatch, which now contains 50\% of synthetic data with known label $\mathbf{t}$, and 50\% of real-world LDR panoramas sampled from the SUN360 database~\cite{xiao2012recognizing}, Google Street View, and training images (where ground truth is ignored) from our real dataset. Unfortunately, applying this domain-adaptated model does not yield satisfying quantitative results (table~\ref{tab:quant-augmentation}), probably because there are not enough unlabelled theta S panoramas. However, we use this model when testing on SUN360 and Google Street View imagery in sec.~\ref{sec:app} with improved results. 

\vspace{-1em}
\subsubsection{Adapting the input panorama}

Aside from augmenting the training data, another option is to adapt the input panorama $\mathbf{x}_\text{LDR}$. For this, we apply different camera models to the JPG files captured by the theta S camera, ranging from: 1) none (``JPGs''); 2) a simple $\gamma = 2.2$ as is commonly done in the literature; 3) a per-channel inverse response function calibrated using a Macbeth color checker (``RF''); and 4) the inverse response function followed by a white balance transformation (``RF+WB''). These options are compared in table~\ref{tab:quant-linearization}. We observe that
linearizing the input data with the ``RF+WB'' model performs best. If such information is unavailable, the model trained on augmented synthetic data (``Tone'') performs best on the input JPG images themselves.

\begin{figure}[!t]
\footnotesize
\centering 
\setlength{\tabcolsep}{1pt}
\begin{tabular}{ccc}
\rotatebox{90}{Background image} & 
\includegraphics[width=.45\linewidth]{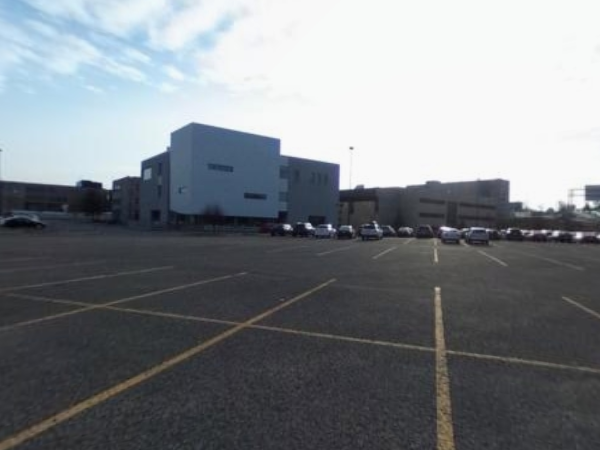}  & 
\includegraphics[width=.45\linewidth]{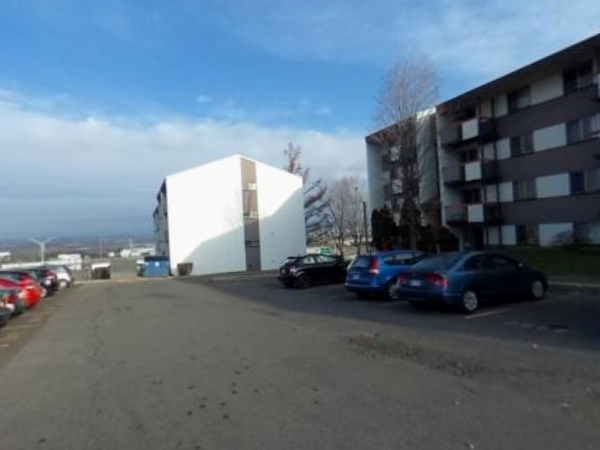} \\
\rotatebox{90}{Render with GT} & 
\includegraphics[width=.45\linewidth]{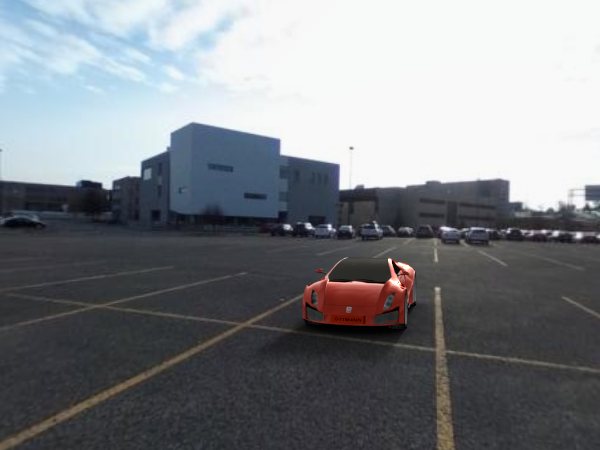}  & 
\includegraphics[width=.45\linewidth]{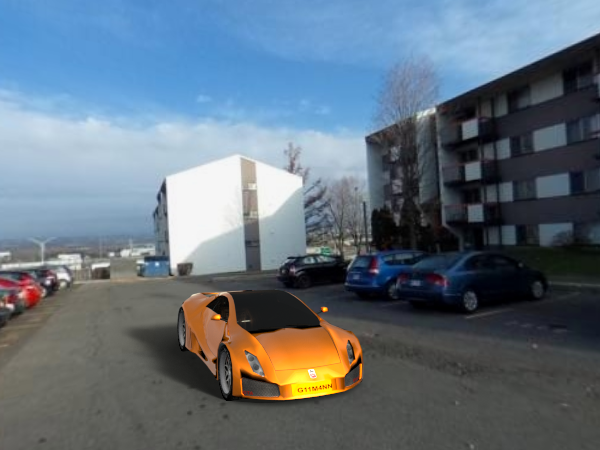} \\ 
\rotatebox{90}{Render with prediction} & 
\includegraphics[width=.45\linewidth]{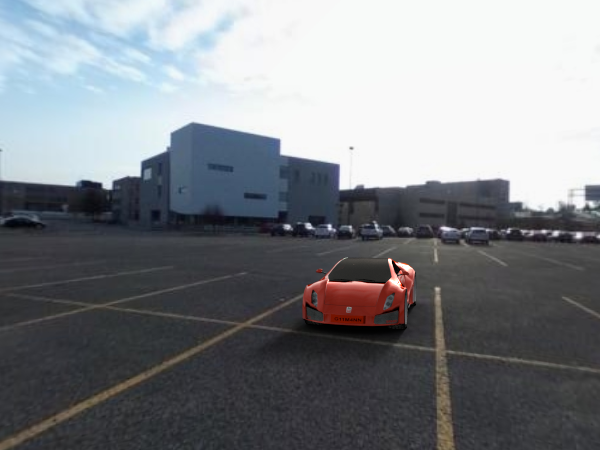} & 
\includegraphics[width=.45\linewidth]{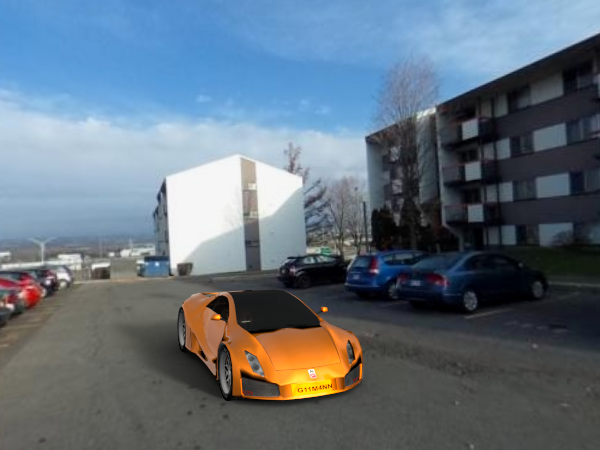} 
\end{tabular}
\vspace{.5em}
\caption{Capturing outdoor light probes with a single shot. A single LDR panorama is shot at the point of object insertion (top). Our approach correctly extrapolates the high dynamic range from the panorama, resulting in a realistic render (last row). Please see additional results on our project page. }
\label{fig:application-theta}
\vspace{-1.5em}
\end{figure}

\vspace{-1em}
\subsubsection{Adapting with ground truth HDR}
\label{subsubsection-finetuning}
We observe that with a small amount of labelled data, the model can be fine-tuned to gain a better performance. We use JPG panoramas from the novel real dataset (8 days of training subset) to fine-tune the best model trained with JPGs (``Tone''), and report the results in the last row of table~\ref{tab:quant-augmentation}. This option far outperforms the others, and can learn to robustly predict HDR from LDR. Fig.~\ref{fig:quant-violin} shows the distribution of errors for each method evaluated in this section. In addition, it also shows qualitative examples corresponding to the 25th, 50th and 75th percentile of errors for the fine-tuned network to illustrate the meaning of these numbers. 

\vspace{-.5em}
\subsection{Comparison with previous work}

Since there are very many options in the inverse tone-mapping (iTMO) literature, we compare against the following set of representative methods: \cite{banterle2007framework,kuo2012content,meylan2006reproduction,rempel-siggraph-07}. We also experimented with the approach of Hold-Geoffroy et al.~\cite{hold-geoffroy-cvpr-17}. They propose to fit a physically-based sky model~\cite{hosek-siggraph-12} to the unsaturated portion of the sky, and extrapolate the sun color via an additional sun model~\cite{hosek-cga-13}. 

Table.~\ref{tab:quant-comparison} shows the errors for each method computed on our test set. For fairness, the parameters of each method (except \cite{hold-geoffroy-cvpr-17} which does not require tuning) are cross-validated on the training set. Our method is the fine-tuned model on the same training set. As table~\ref{tab:quant-comparison} shows, the existing iTMO methods fail at this task. This is probably due to the fact that they were not designed to work with the extremely high dynamic range of outdoor lighting. We found the method of Hold-Geoffroy et al.~\cite{hold-geoffroy-cvpr-17} to consistently overshoot in estimating the sun intensity.


\begin{figure}[!t]
\footnotesize
\centering 
\setlength{\tabcolsep}{1pt}
\begin{tabular}{ccc}
\rotatebox{90}{Background image} & 
\includegraphics[width=.45\linewidth]{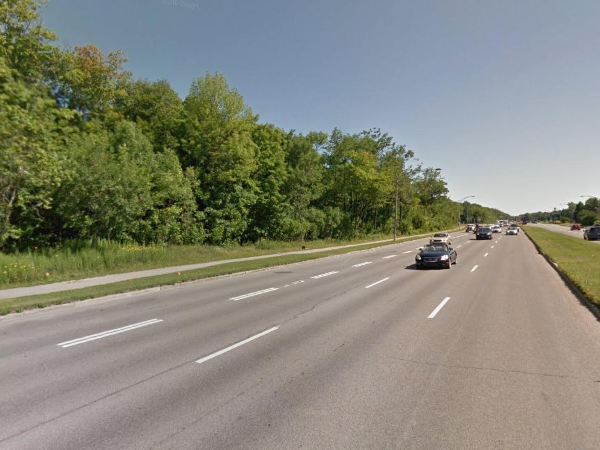} & 
\includegraphics[width=.45\linewidth]{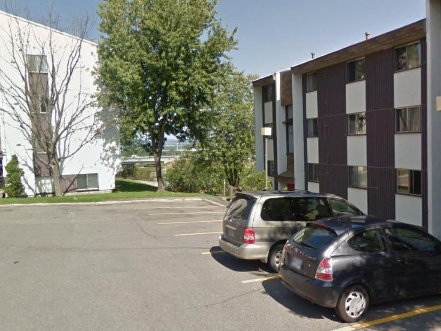} \\

\rotatebox{90}{Render with prediction} & 
\includegraphics[width=.45\linewidth]{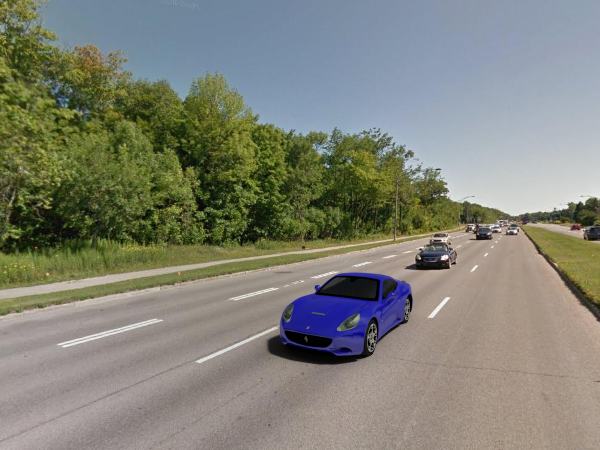} &
\includegraphics[width=.45\linewidth]{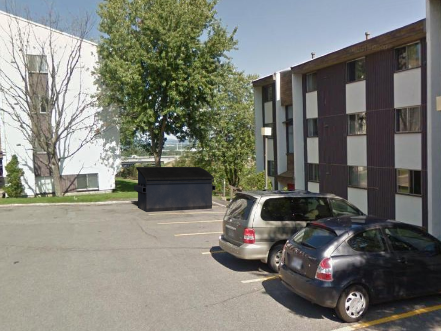} \\
\end{tabular}
\vspace{.5em}
\caption{Relighting in a Google Street View image. Every Google Street View image is a LDR spherical panorama. Our method can be applied to this dataset to automatically estimate plausible HDR data and be used, for example, to visualize virtual objects in real scenes. The top row shows a cropped regular image from the panorama, and the bottom shows a virtual object relit with the HDR panorama predicted from our network. Our method can realistically extrapolate high dynamic range even on uncalibrated cameras. Please see additional results on our project page. }
\label{fig:application-google}
\vspace{-2em}
\end{figure}

\begin{figure*}[!t]
\centering
\footnotesize
\setlength{\tabcolsep}{1pt}
\newcolumntype{R}{c@{\extracolsep{0pt}}c@{\extracolsep{0pt}}}%
\begin{tabular}{cRRR}
Target & NN1 & NN2 & NN3 & NN4 & NN5\\
\raisebox{.5em}{\rotatebox{90}{bright, $75^\circ$}} & 
\includegraphics[height=1.62cm]{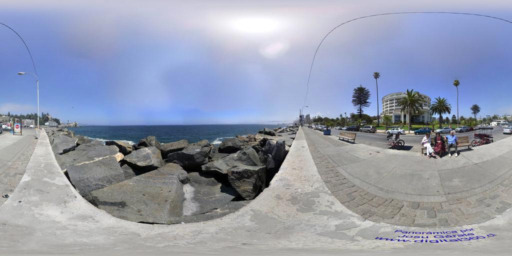} &
\includegraphics[height=1.62cm]{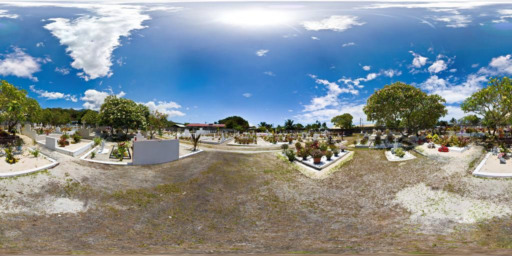} &
\includegraphics[height=1.62cm]{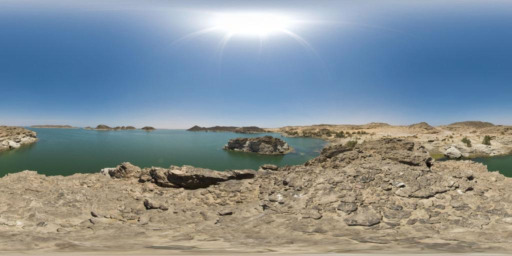} &
\includegraphics[height=1.62cm]{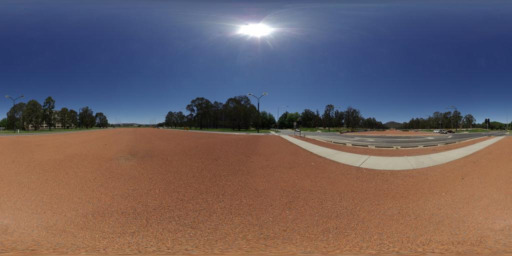} &
\includegraphics[height=1.62cm]{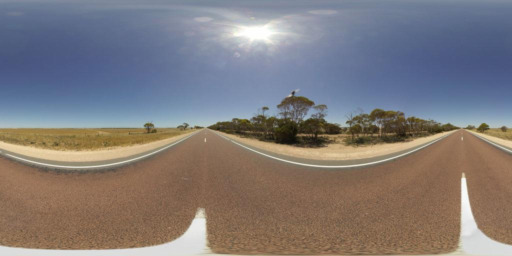} \\

\raisebox{1em}{\rotatebox{90}{dim, $75^\circ$}}& 
\includegraphics[height=1.62cm]{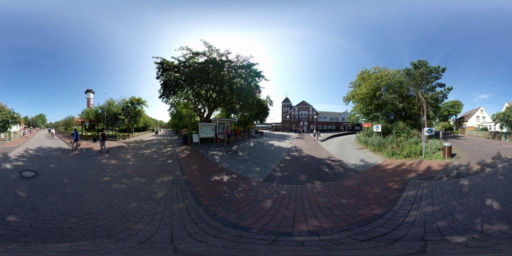} &
\includegraphics[height=1.62cm]{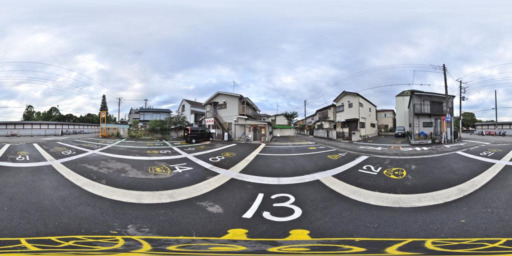} &
\includegraphics[height=1.62cm]{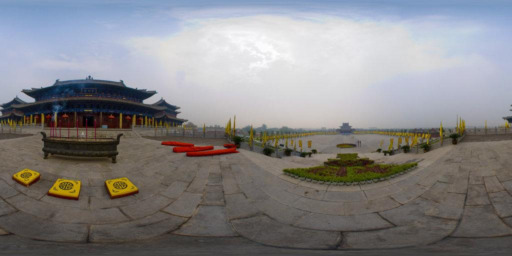} &
\includegraphics[height=1.62cm]{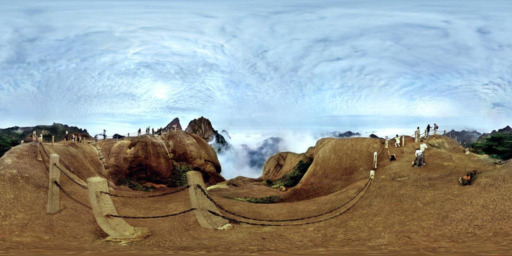} &
\includegraphics[height=1.62cm]{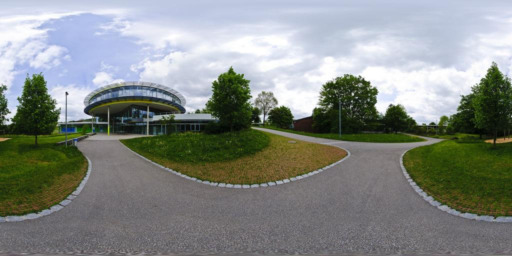} \\
\raisebox{.5em}{\rotatebox{90}{bright, $20^\circ$}}& 
\includegraphics[height=1.62cm]{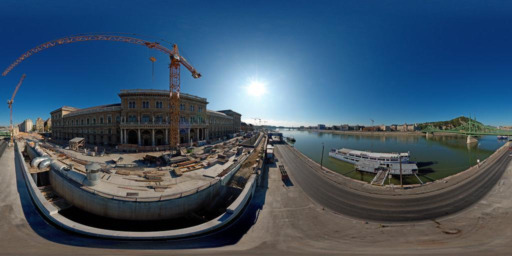} &
\includegraphics[height=1.62cm]{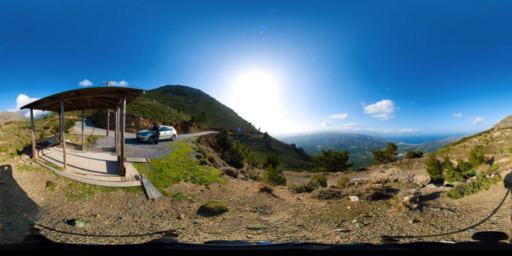} &
\includegraphics[height=1.62cm]{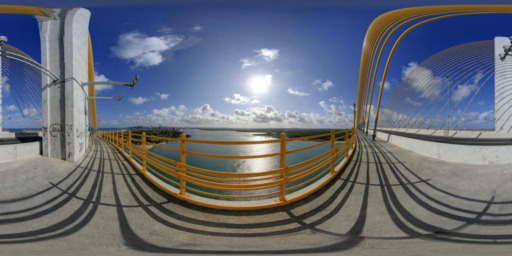} &
\includegraphics[height=1.62cm]{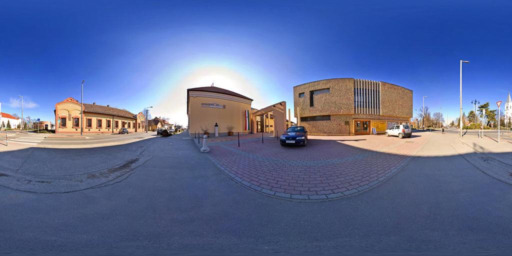} &
\includegraphics[height=1.62cm]{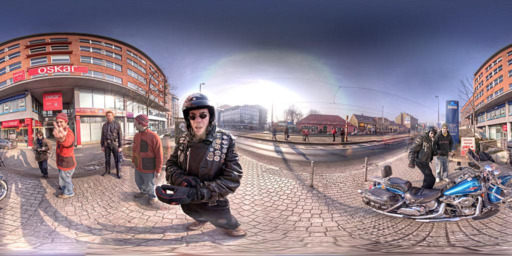} \\

\raisebox{1em}{\rotatebox{90}{dim, $20^\circ$}}& 
\includegraphics[height=1.62cm]{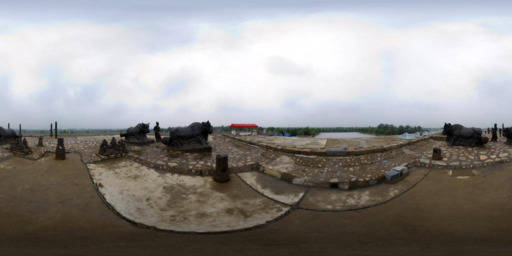} &
\includegraphics[height=1.62cm]{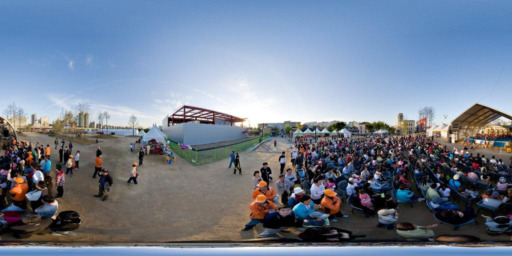} &
\includegraphics[height=1.62cm]{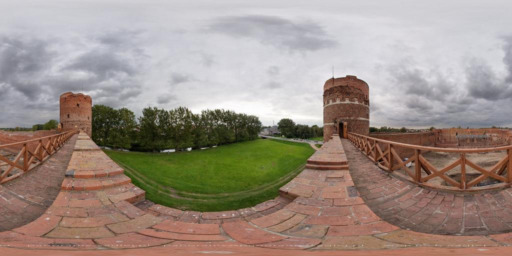} &
\includegraphics[height=1.62cm]{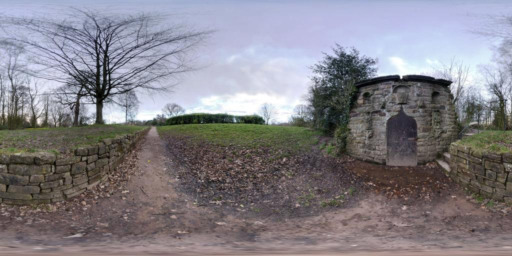} &
\includegraphics[height=1.62cm]{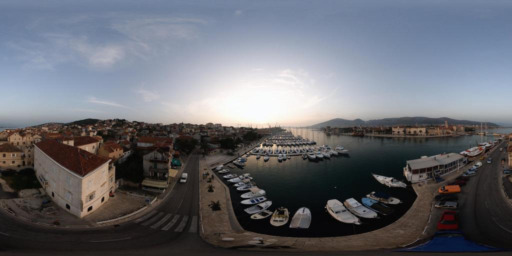} \\
\end{tabular}
\vspace{.5em}
\caption{Illumination-based image matching from the SUN360 database~\cite{xiao2012recognizing}. We retrieve images based on two target parameters: the sun intensity (bright or dim) and elevation (in degrees). Matching LDR panoramas to these parameters would not be possible, so images are retrieved based on the estimates given by our network (the sun intensity is simply the intensity of the brightest pixel in the HDR panorama). Here, a bright (dim) value is given as the 75th (25th) percentile sun intensity over the entire dataset.}
\label{fig:application-sun360}
\vspace{-1.2em}
\end{figure*}

\vspace{-.5em}
\section{Applications}
\vspace{-.2em}
\label{sec:app}
Extrapolating accurate high dynamic range from a single, low dynamic range panorama gives rise to several interesting applications. In this section, we present three different ways of using our network in practical scenarios. 

\subsection{Single shot outdoor light probe}

We can use our method to simplify the process of capturing outdoor light probes. Because of its extremely high dynamic range, capturing outdoor light requires a carefully calibrated setup~\cite{stumpfel-afrigraph-04} or specially-designed light probes~\cite{debevec-st-12}. Instead, one could simply take an LDR panorama with an off-the-shelf, point-and-shoot $360^\circ$ camera such as the Ricoh Theta S, and extrapolate the HDR using our network. 

This is demonstrated in fig.~\ref{fig:application-theta}. In this example, a novice user shot an LDR panorama with a Theta S at the location where the object is to be inserted. From this LDR panorama, the prediction from our ``Fine-tuning'' network is used as a light probe for image-based lighting to insert virtual objects in the photograph. The image rendered with our prediction produces a plausible rendering result.

\subsection{Visualization in Google Street View imagery}
\label{sec:street-view}

The Google Street View dataset is a huge source of LDR panoramas, captured all over the world by cars equipped with omnidirectional cameras. We leverage our approach to extrapolate HDR from this dataset, and show that the resulting panoramas can be used for image-based lighting in fig.~\ref{fig:application-google}. In this case, we use the domain-adapted version of the network from sec.~\ref{sec:quantitative-evaluation} as it qualitatively yields better results than the network fine-tuned on the theta S data.

\subsection{Image matching in the SUN360 dataset}

We can also use our method for matching panoramas based on intuitive illumination parameters. For instance, we show an example of browsing the SUN360 database~\cite{xiao2012recognizing} based on sun elevation and intensity in fig.~\ref{fig:application-sun360}. For this, we first run our network on all outdoor LDR panoramas in the database. Then, we compute the sun intensity as being the intensity of the brightest pixel in the estimated HDR panoramas. Finally, a target set of parameters is specified (e.g. bright sun at $75^\circ$ elevation as in the first row of fig.~\ref{fig:application-sun360}), and the best images can efficiently be retrieved using nearest neighbor matching. Note that this would be very hard to do without our method, as the sun is always saturated in outdoor panoramas. As with sec.~\ref{sec:street-view}, the domain-adapted version of the network from sec.~\ref{sec:quantitative-evaluation} is used here as well. 

\vspace{-.2em}

\section{Conclusion}

We present a full end-to-end learning approach to estimate the extremely high dynamic range of outdoor lighting from a single, LDR $360^\circ$ panorama. Our main insight is to exploit a large dataset of synthetic data composed of a realistic virtual city model, lit with real world HDR sky light probes~\cite{hdrdb}, to train a deep convolutional autoencoder. The resulting network is evaluated on a wide range of experiments on synthetic data, as well as on a novel dataset of real LDR panoramas and corresponding HDR ground truth. The applicability of the approach is also demonstrated on three novel applications. 

Despite its success, our approach is not without limitations. First, we note a certain sensitivity to the tonemapping function of the input LDR. Our qualitative experiments demonstrate that domain adaptation helps in adjusting to the wide variability in camera response functions and image post-processing operations applied to panoramas in the SUN360 dataset~\cite{xiao2012recognizing}, but these are complicated by the fact that no ground truth is available. A second limitation is that our approach is limited to outdoor scenes and the sun, when visible, needs to be centered in the panorama. While the sun is typically relatively easy to detect in LDR panoramas, the simple sun detection method of~\cite{hold-geoffroy-cvpr-17} may fail, resulting in unlikely results. A possible extension of this work could be the inclusion of sun azimuth estimation from the FC-64 latent layer. Finally, the resolution of the output is limited at $64 \times 128$, which, while sufficient for relighting applications, cannot extrapolate the HDR information in a full-resolution LDR background image. A potential way to leverage high resolution images is applying a fully convolution network~\cite{long2015fully} by converting all fully-connected layers to convolutions. Future work includes the adaptation of the network to learn high resolution HDR textures from limited field-of-view LDR images. 

\vspace{-.5em}
\section*{Acknowledgement}
We thank Mathieu Garon, Yannick Hold-Geoffroy, Henrique Weber, and Maxime Tremblay for helpful discussions and comments, and Jean-Michel Fortin for his help in capturing images for the novel dataset. This work was partially supported by the FRQNT New Researcher Grant 2016NC189939, the NSERC Discovery Grant RGPIN-2014-05314, and the REPARTI Strategic Network. We gratefully acknowledge the support of Nvidia with the donation of the GPUs used for this research.

{\small
\bibliographystyle{ieee}
\bibliography{mainbib}
}

\end{document}